\documentclass[10pt, a4paper, twocolumn, teaser, showabstract]{naverlabseurope}

\usepackage{multirow}
\usepackage{xspace}
\usepackage{tikz,pgfplots}
\usepgfplotslibrary{groupplots}
\pgfplotsset{compat=1.16}
\pgfplotsset{set layers}
\usepackage{enumitem}
\setlist[itemize]{topsep=0pt, itemsep=0pt, label=\textbullet}
\usepackage{stfloats}
\usepackage[capitalise,nameinlink]{cleveref}

\graphicspath{{tex/figures/}}

\newcounter{rownumbers}
\newcounter{rownumbersft}

\crefname{section}{Sec.}{Secs.}
\Crefname{section}{Sec.}{Secs.}
\Crefname{table}{Tab.}{Tabs.}
\crefname{table}{Tab.}{Tabs.}
\Crefname{figure}{Fig.}{Figs.}
\crefname{figure}{Fig.}{Figs.}
\Crefname{equation}{Eq.}{Eqs.}
\crefname{equation}{Eq.}{Eqs.}
\crefname{appendix}{Appendix}{Appendices}

\makeatletter
\newcommand{\CapitalizeFirst}[1]{
  \expandafter\@CapitalizeFirst\expandafter{#1}
}
\def\@CapitalizeFirst#1{\MakeUppercase{\@car#1\@nil}\@cdr#1\@nil}
\makeatother

\newcommand{\ours}{SPAR3S\xspace}

\makeatletter
\DeclareRobustCommand\onedot{\futurelet\@let@token\@onedot}
\def\@onedot{\ifx\@let@token.\else.\null\fi\xspace}
\def\eg{\emph{e.g}\onedot} 
\def\ie{\emph{i.e}\onedot}

\makeatother

\renewcommand{\paragraph}[1]{\vspace{1pt}\noindent\textbf{#1}}

\definecolor{tabdefault}{gray}{0.8}
\definecolor{lightgray}{gray}{0.93}

\definecolor{tabsecond}{rgb}{0.8, 1, 0.8}
\definecolor{tabthird}{rgb}{0.88, 1, 0.88}
\definecolor{tabfirst}{rgb}{0.5, 1, 0.5}
\definecolor{tabtop}{rgb}{0.2, 1, 0.2}
\definecolor{tablast}{rgb}{1, 0.5, 0.5}
\definecolor{tablastest}{rgb}{1, 0.2, 0.2}
\definecolor{tabsecondlast}{rgb}{1, 0.8, 0.8}

\definecolor{teal}{RGB}{41,120,108}

\title{Sparse auto-regressive modeling for scene generation from multi-view images}
\titlerunning{\ours: sparse auto-regressive modeling for scene generation from multi-view images}

\correspondingauthor{[thomas.lucas,maxime.pietrantoni]@naverlabs.com}

\authors{%
Thomas Lucas$^{1,\star}$,
Maxime Pietrantoni$^{1,\star}$,
Philippe Weinzaepfel$^{1}$,
Wonjune Cho$^{2}$,
Bardienus Pieter Duisterhof$^{3}$,
Vincent Leroy$^{1}$,
Jerome Revaud$^{1}$}
\affiliations{$^{1}$NAVER LABS Europe, France \quad $^{2}$NAVER LABS, South Korea \quad $^{3}$Carnegie Mellon University, USA}
\contributions{$^{\star}$equal contribution \\[4pt]
\textit{European Conference on Computer Vision (ECCV) 2026}}
\website{}
\websiteref{}

\teaserfig{\includegraphics[width=0.95\textwidth]{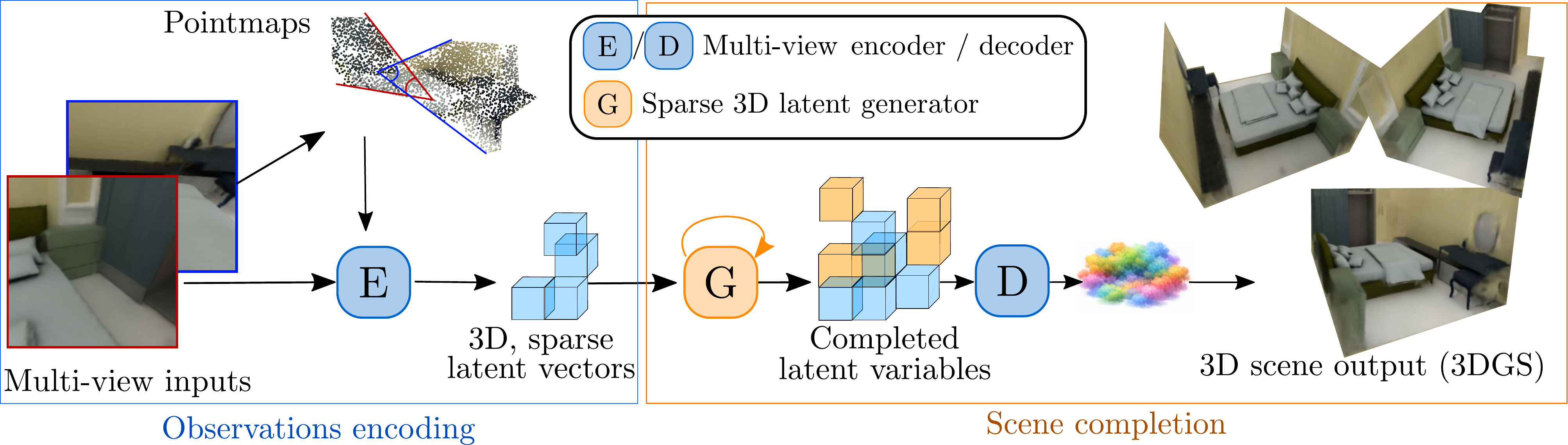}}
\teasercaption{\textbf{Overview of \ours.} An encoder extracts voxel-aligned conditioning 3D latent tokens (in blue) from sparse (\eg 2) input views. This encoder is trained together with a decoder that outputs 3D Gaussians that can be rendered via splatting. The partially filled 3D latent voxel grids are iteratively expanded by a 3D masked auto-regressive transformer that jointly predicts the occupancy of the remaining voxel grids, and their corresponding latent tokens (in yellow). The generated latent tokens can be decoded into 3D Gaussians using the scene decoder.\label{fig:teaser}}

\begin{abstract}
Generating complete 3D scenes from sparse, unconstrained views is a fundamental challenge in 3D vision which requires reasoning beyond observed content while remaining computationally tractable.
Existing feed-forward reconstruction methods are inherently limited to content visible in the input images, while 3D generative modeling is hindered by the high computational cost of dense volumetric representations and the scarcity of large-scale 3D supervision.
We introduce \ours, a sparse voxel-aligned 3D latent generative model for conditional scene completion without requiring ground-truth 3D data for supervision. 
Our key insight is to formulate 3D scene generation in a structured, compact, voxel-aligned 3D latent space where only occupied voxels are represented. We learn this sparse latent space directly from multi-view images using photometric supervision via differentiable 3D Gaussian Splatting.
Given a partial set of observed voxels encoded from sparse input views, scene completion reduces to predicting the missing latent tokens and their spatial support within the voxel grid. To this end, we train a masked autoregressive transformer that jointly models voxel occupancy and latent token values, enabling efficient and spatially consistent generation of unseen regions.
We demonstrate the effectiveness of our method on synthetic indoor scenes, achieving higher novel-view quality than prior work.
We further validate its generalization on RealEstate10k, highlighting its applicability to real-world data.
\end{abstract}

\begin{document}

\maketitle

\section{Introduction}
\label{sec:introduction}

Recent advances in novel view synthesis, \eg 3D Gaussian Splatting (3DGS)~\cite{kerbl20233dgaussiansplatting}, have enabled real-time, photorealistic rendering of 3D scenes, optimized through differentiable photometric supervision. In parallel, feed-forward multi-view stereo and pointmap regression methods such as DUSt3R~\cite{wang2024dust3r} have demonstrated that 3D geometry can be recovered directly from images without iterative optimization. 
Combined, these paradigms lead to \emph{pixel-aligned} 3D Gaussian regression from sparse views~\cite{charatan2024pixelsplat,xu2024depthsplat,hen2024mvsplat}. 
However, pixel-aligned predictions are fundamentally limited to the content visible in the input images.

To complete scenes, a recent line of work combines multi-view diffusion models to generate 2D novel content with pixel-aligned Gaussian Splats regression further lifted in 3D~\cite{henderson2024sampling,szymanowicz2025bolt3d,bahmani2025lyra}. 
Such approaches require known cameras to generate new content or strong assumptions about them, \eg with object-centric tasks or 2D panoramic image generation. This leads to a chicken-and-egg problem, where sampling plausible cameras requires knowing the scene and vice-versa. 
Additionally, multi-view diffusion models lack geometric consistency across novel views because of a lack of explicit 3D representation. 
By contrast, generating content directly in 3D does not require a known set of novel camera poses beforehand, and inherently enforces geometric consistency while naturally handling unconstrained camera positions.
Therefore, in this work we aim to generate unseen content in the form of 3DGS representations directly in 3D from sparse unconstrained input views.  

Generating complete 3D scenes from sparse, unconstrained views requires reasoning beyond observations, and modeling the distribution of plausible unseen geometry and appearance. 
To do so, we propose to lift the problem into a \emph{voxel-aligned 3D latent space}. Concretely, a first model learns to map multi-view observations of a scene to voxel-aligned 3D latent variables.
Then, a generative model is trained to learn a distribution over the 3D latent space;
when parts of a 3D scene are not observed from a set of multi-view inputs, corresponding voxels can be sampled from the generative model to complete the 3D scene.
The completed latent representation is then decoded into a 3DGS representation, see Fig.~\ref{fig:teaser} for an overview.  
Designing such generative models in 3D is challenging for two main reasons. First, dense volumetric representations scale cubically with spatial resolution, making high-resolution generative modeling computationally prohibitive. Second, large-scale curated 3D supervision is scarce unlike with images or text, limiting the applicability of fully-supervised 3D generative learning. 

Our key idea to address scalability is to only materialize occupied voxels in our structured voxel-aligned latent space. Our encoder-decoder architecture maps arbitrary sets of multi-view input images to a  \emph{sparse} and \emph{compact} voxel-based 3D latent representation, which is decoded into a 3DGS scene with voxel-aligned Gaussians.
To model a distribution over such latent representations, we introduce a \emph{sparse autoregressive transformer} operating over sequences of 3D tokens. Scene completion thus reduces to predicting both voxel occupancy and latent token features. 
By exploiting sparsity, spatial compression, and structured 3D token orderings, our approach is efficient while preserving volumetric coherence. 
To address raw 3D data scarcity, our sparse latent space is learned directly from multi-view images using \emph{photometric supervision}, via differentiable 3D Gaussian Splatting. Importantly, this learning process does not require any ground-truth 3D supervision.

We validate our proposed 3D generative paradigm, called \ours 
on synthetic indoor scenes, where it achieves improved novel-view synthesis quality compared to prior feed-forward 3DGS regression methods as well as 3D generative methods. We further demonstrate generalization to real-world data on RealEstate10K~\cite{zhou2018stereo}. 

Our contributions may be summarized as follows:
\begin{itemize}
\item We introduce a sparse voxel-aligned 3D latent space learned from multi-view photometric supervision (Section~\ref{sec:ae}). 
\item We propose an occupancy-aware masked autoregressive transformer for conditional scene completion which operates in this latent space (Section~\ref{sec:painter_carver}).
\item We demonstrate improved novel-view synthesis quality over prior feed-forward 3DGS generative and deterministic approaches, with validation on both synthetic and real-world datasets (Section~\ref{sec:experiments}).
\end{itemize}

\section{Related work}
\label{sec:related}

\paragraph{Observed 3D scene reconstruction} aims to regress observed scene geometry from multi-view inputs. 
The resulting scene representations only cover the parts of the scene that are observed in the input images.
We rely on such partial representations, but additionally aim to complete unobserved parts of the scene with a generative model and capture visual content together with geometry. 
Recent methods for scene reconstruction like DUSt3R \cite{wang2024dust3r} and MASt3R \cite{leroy2024grounding} represent a scene by predicting a dense per-pixel 3D point map $P(u,v)\in\mathbb{R}^3$ mapping each image pixel directly to its 3D coordinate in camera space.
We use such pointmaps, associated to a set of multi-view inputs, to  initialize a sparse voxel aligned latent space and remove voxels that do not contain anything when encoding targets.
 Conveniently, these methods have been extended to accommodate more than 2 views, either by introducing some scene-level memory \cite{wang20243d,cabon2025must3r,wang2025continuous} or by scaling attention layers to a larger number of tokens from multi-views \cite{wang2025vggt}. 

\paragraph{Feed-Forward 3D Gaussian splatting.} 
Feed-forward 3D Gaussians models have been introduced to reconstruct a scene from a sparse number of input views\cite{szymanowicz2024splatter,charatan2024pixelsplat,hen2024mvsplat,szymanowicz2024flash3d,xu2024depthsplat,tang2024lgm,xu2024grm,zhang2024gs,liu2025monosplat}. The base principle lies in regressing 3D Gaussians parameters from a pixel-aligned feature map \cite{szymanowicz2024splatter}. PixelSplat~\cite{charatan2024pixelsplat} predicts 3D Gaussians parameters from a pair of images by relying on epipolar cues while MVSplat \cite{hen2024mvsplat} aggregates multi-view information in a cost volume before predicting Gaussians, implicitly learning to match and triangulate. To overcome the ambiguity of reconstructing texture-less regions, \cite{xu2024depthsplat,szymanowicz2024flash3d,liu2025monosplat} further introduce monocular depth priors as regularization. Instead of using such priors, \cite{tang2024lgm,xu2024grm,zhang2024gs} directly predict a set of 3D Gaussians from images using a large transformer model, at the cost of a significant increase in training budget. 
These methods lack structured 3D representations, which limits geometric consistency across views as well as their ability to handle very sparse inputs and extreme viewpoint variations. In contrast, our model is designed around a latent sparse 3D space to handle these difficult cases.
Furthermore, standard feed-forward 3DGS approaches are deterministic and fail to reconstruct ambiguous regions that may be occluded and not observed in the input views. We now present several distinct lines of work that introduce generative modeling to address this challenge. 

\paragraph{Multi-view diffusion over image space.}
The strong generative priors contained in image and video diffusion models have been used to generate novel views conditioned on camera poses \cite{gao2024cat3d,liang2025wonderland,ma2025you,zhou2025stable}. To improve geometric consistency of the generated images, some approaches \cite{liu2024reconx,yu2024viewcrafter,chen2025scenecompleter} directly condition the denoising process on strong 3D scene priors \cite{wang2024dust3r}. 
In contrast to our task, these methods do not produce an explicit 3D scene representation as output, but rather sets of novel views. Thus, generated content cannot be rendered in real-time in 3D, and the procedure needs to be re-run for each additional set of views. Another drawback is that there is no guaranteed spatial coherence between views, inducing flickering and geometric drift.

\paragraph{Iterative 3DGS optimization using diffusion priors.}
In \cite{tang2023dreamgaussian,li2023gaussiandiffusion} a Score Distillation Sampling loss is introduced to leverage 2D diffusion model priors and directly optimize a 3DGS scene representation. In addition, \cite{yi2024gaussiandreamer} initialize the Gaussians with a diffusion model to obtain a rough initial geometry which facilitate downstream convergence. Any scene reconstruction is therefore based on iterative optimization and takes a significant amount of time.

\paragraph{2D generation with 3DGS prediction.}
Given a set of posed conditioning images and a set of novel camera poses, this class of methods outputs a set of 3DGS parameters for each novel camera pose by using 2D generative models.
In the context of object-centric reconstruction, \cite{meng2025zero,xiang2025repurposing} finetune a StableDiffusion model \cite{rombach2022stablediffusion} to output Gaussian parameters and
\cite{szymanowicz2025bolt3d, bahmani2025lyra} optimize a 3DGS prediction head on top of a pretrained controlled video diffusion model. The head predicts pixel-wise 3DGS parameters for each novel view sampled from the diffusion model.
In \cite{henderson2024sampling,lin2025diffsplat} a 2D encoder-decoder is trained to encode images in a pixel-aligned latent space that can be further decoded into pixel-aligned Gaussians. In a second stage they optimize a multi-view diffusion model in this latent space. 
Such approaches require \emph{known camera} as input to sample novel content in 2D, or alternatively need to make simplifying assumptions such as object-centric targets or 2D panoramic scenes to easily sample cameras.
In contrast, we tackle the more general problem of generating content directly in 3D, without requiring known camera poses or making simplifying assumptions about the scene.

\paragraph{3D Generative feed-forward 3DGS.}
The following methods directly apply diffusion in 3D. 
In \cite{peng2025lesson} a 3D diffusion model is optimized to sample 3DGS, but requires a teacher model and ground-truth splats for training; this limits scalability as such data requires dense sets of views and a slow optimization procedure per scene.
Closer to our work, \cite{wewer24latentsplat} adopts a VAE formulation with variational Gaussians that encode uncertainty in a latent space, and these can be rendered and decoded in 2D to allow for easy photometric optimization. 
\cite{liao2025complete} instead applies diffusion in a 3D latent space to model the conditional distribution of latents that can be decoded to 3D Gaussian splats based on conditioning views. In \cite{chen2024mvsplat360} novel images are sampled with diffusion conditioned on latents rendered from a unified 3D representations. 
These 3D diffusion approaches are used to sample a fixed set of Gaussian parameters which limits the extent of the generated scene and scalability. In contrast, we sample latents from a sparse voxel grid and apply an autoregressive model which provides greater flexibility and representational power. 
More similar to our approach, SCube~\cite{scube} and XCube~\cite{xcube} both sample occupancy grids in a voxelized latent space, but with generative formulations that do not encode appearance and geometry jointly in a latent space.

\begin{figure*}[t]
  \centering
  \includegraphics[width=0.9\linewidth]{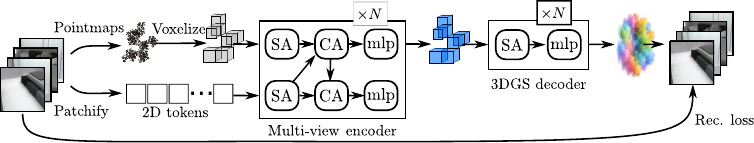}
  \caption{\textbf{Learning a 3D latent space from multi-view input data.} Multi-view inputs are patchified and used to initialize voxel-aligned tokens from pointmaps. A bidirectional cross-attention block updates 2D and 3D tokens. The final 3D latent representation is decoded into a 3D scene represented as a set of Gaussian splats. The model is trained to reconstruct the inputs via differentiable rendering. No explicit 3D scene ground truth is required, yet this produces 3D representations.
  }
  \label{fig:main1}
\end{figure*}

\paragraph{Autoregressive latent modeling.}
Our approach is based on auto-regressively predicting sparse 3D latent tokens, and borrows from  autoregressive models primarily developed for images.
In particular, discrete latent approaches 
such as VQ-VAE\cite{van2017neural} and its hierarchical extension VQ-VAE-2\cite{razavi2019vqvae2} represent images using compact codebooks and train autoregressive priors over these quantized latents. We follow the same overall paradigm, but adapt it to 3D.
Recently, masked and autoregressive~\cite{chang2022maskgit,li2022mage,li2024autoregressive}. transformers have emerged as flexible alternatives to raster-order PixelCNN-style\cite{oord2016pixelcnn} models: they allow inference over tokens in arbitrary order. 
We extend this flexibility to sparse 3D voxel grids, leveraging arbitrary token orderings to operate directly on sparse voxel representations. 
Other applications of auto-regressive models to 3D  include \cite{leumami}, which uses a multiview 2D transformer, and \cite{chen2025mar} adapts it to object-centric generation. In contrast, we directly operate in a 3D voxel-aligned latent space which enforces geometric consistency across views and does not require simplifying assumptions about camera distribution.

\section{3D latent representation from multi-view data} 
\label{sec:ae}
Our scene encoder-decoder model follows the general framework of auto-encoders: it encodes input views into a latent representation, and is trained to reconstruct these inputs by decoding the latent variables.
However, in our case the inputs and outputs are only a \emph{proxy} to the quantity that we seek to model, which is 3D scenes represented via 3DGS.
To achieve this, the flow of information in our encoder-decoder is \emph{constrained to go through a single 3DGS representation per scene}, from which all input views should be reconstructed.
This constraint can be seen as a type of bottleneck in the general auto-encoder framework as all the information in the inputs has to be compressed into a single 3D representation. 
Additionally, we require that this 3DGS representation should be computed from a \emph{single, voxel aligned} 3D latent representation. 
With this construction it is possible to generate scene content by first generating a 3D latent representation, then decoding it with the scene decoder.

Let  $\mathcal{I} {=} \{ I_i \}_i$ be a set of multi-view input RGB images and $E_\theta$ an encoder that maps multi-view inputs into a latent 3D representation $z_{\text{3D}}$.
Also let $D_{\phi}$ a decoder that maps such a 3D latent variable to a 3DGS scene representation denoted $G$,
and let $\mathcal{R}$ a rendering operation that reconstructs $\hat{\mathcal{I}}$ from $G$ via differentiable rendering.
Our scene encoder-decoder can be summarized as: 
\begin{equation} 
Z_{\text{3D}} = E_\theta(\mathcal{I}), ~~\: G = D_\phi(Z_\text{3D}), ~~\: \hat{\mathcal{I}} = \mathcal{R}(G).
\end{equation}
$E_\theta$ and $D_\phi$ are optimized with a photometric reconstruction loss; see Fig.\ref{fig:main1} for an overview.
The two 3D representations $Z_{\text{3D}}$ and $G$ are produced as a byproduct of this auto-encoding problem, yet modeling these quantities is the real purpose of our latent generative model. 

\begin{figure*}[t]
  \centering
  \includegraphics[width=0.9\linewidth]{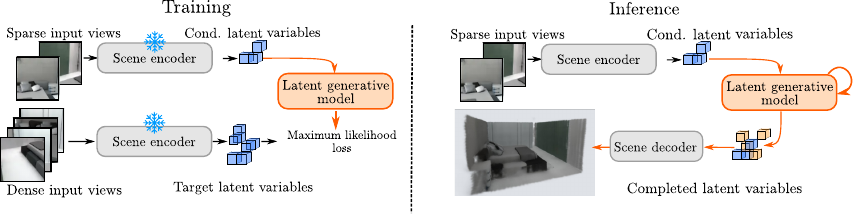}
  \caption{\textbf{Learning a latent generative model.} For training, a dense set of multi-view inputs is aggregated into a 3D latent representation used at target for the generative model. This target is predicted conditionally on an other latent representation, encoded from sparse observed views. 
  Inference starts from the observed latent variables which are iteratively expanded to complete the scene in latent space before decoding it to 3DGS.}
  \label{fig:main2}
\end{figure*}

\noindent \paragraph{Learning a sparse compact 3D latent space.}
In practice, processing a full 3D token grid is computationally prohibitive unless voxel resolution is very low. Hence, most of the key design choices made for our model aim at addressing this bottleneck.
First, we design our auto-encoder such that it operates on a sparse 3D representation: only tokens that correspond to voxels \emph{with content} (ie. non empty) will be processed by the scene encoder. 
Second, we build a hierarchical encoder-decoder such that the latent space is more compact than the scene. The encoder progressively downsamples 3D voxels until the bottleneck. The decoder then upsamples this latent space back to the original voxel resolution.
Both of these aspects drastically alleviates the computational load and allow us to derive a model operating fully in 3D.

\noindent \paragraph{Sparse 3D voxel initialization.} Our encoder model processes sets of multi-view input RGB images  \( \mathcal{I} {=} \{ I_i \}_i \). First, pointmaps\cite{wang2024dust3r}  $\{P_i\}_i$ are estimated from a scene with \emph{unconstrained} viewpoints, here using \cite{duisterhof2025mastrsfm}. The union of pointmaps \( \mathcal{P} = \bigcup_{i} P_i \)  is rescaled and translated to fit a predefined volume $\mathcal{V}$ with a transformation $T$, such that \( T(\mathcal{P}) \subseteq \mathcal{V} = [0, 1]^3 \). Then 3D latent queries $z_{\text{3D}}^0$ are initialized using the pointmaps to filter out empty voxels; the 2D multi-view inputs are patchified and tokenized into 2D tokens denoted $z_\text{2D}^0$.

\noindent \paragraph{Sparse attention.} A stack of $L$ bidirectional cross-attention blocks, denoted CA and composed of a self-attention layer, a cross-attention layer and an MLP,  is used to update both 3D and 2D tokens: \begin{equation}\left(z_{\text{3D}}^{k+1}, z_{\text{2D}}^{k+1}\right) = \left(CA(z_\text{3D}^k, z_\text{2D}^k), CA(z_\text{3D}^{k+1}, z_\text{2D}^{k})\right).\end{equation}
The 2D tokens are then discarded, and $z_\text{3D}^L$ is decoded with a transformer model $D_\phi$, into a set of \( M \) 3D Gaussians per occupied voxels.
Denoting \(V_{occ}\) the set of occupied voxels and $G$ the set of parameters of one 3D Gaussian splat, we have:
\begin{equation}
\{G_i\}_{i=1}^{M\times |V_\text{occ}|}= D_\phi(Z).
\end{equation}
Throughout these computations, empty voxels are never considered.

\paragraph{Token resampling and sparsification.} The 3D voxel tokens are downsampled through the layers of the encoder into the bottleneck latent space and upsampled through the layers of the decoder. When downsampling, tokens within a local neighborhood are aggregated. When upsampling, a mechanism to maintain spatial sparsity is needed. We use a binary cross-entropy (BCE$\uparrow$) head over upsampled voxel positions to predict voxel occupancy before computing the upsampled token values. During training, the ground-truth (GT) occupancy mask is substituted, augmented with false-positive noise injection to improve the model's robustness to classification errors at inference time. This can be formalized as:
\begin{equation}
\begin{aligned}
z_{\text{3D}}^{\text{up}} &= \text{Upsample}\left(z_{\text{3D}}, \mathcal{M}\right), \quad \text{where} \\
\mathcal{M} &= \begin{cases} \text{BCE$\uparrow$}(z_{\text{3D}}) & \text{at inference} \\ \text{GT} \cup \epsilon_{\text{FP}} & \text{at training.} \end{cases}
\end{aligned}
\end{equation}
where $\epsilon_{\text{FP}}$ represents the sampled false-positive noise distribution.

\paragraph{Geometry-guided attention.}
To facilitate learning, attention maps in the CA blocks incorporate geometry-aware correspondence information, derived from the pointmap $P$ projections onto patches and a bincount between patches and voxels. 
Attention logits are computed as a weighted average between learned attention logits from voxel \( v \) to patch \( p \), 
and    
bincount attention scores:
\begin{equation}
\begin{split}
 A^{\text{guided}}_{v, p} =& \alpha \cdot \sigma(A^{\text{learned}}_{v, :})_p + (1 - \alpha) \cdot \sigma(A^{bincount}_{v, :})_p,
\end{split}
\end{equation}
where
\( \sigma(\cdot)\) denotes a softmax
  and \( \alpha \in [0, 1] \).

\noindent \paragraph{Photometric supervision.} Input images are reconstructed by projecting \(D_\phi(Z)\) onto corresponding cameras, using camera parameters $C$ obtained together with the pointmaps \(P\), with a differentiable Gaussian splatting operation \( \mathcal{R} \) \cite{kerbl20233dgaussiansplatting}. The model is trained end-to-end with an $L_2$ reconstruction loss. Following \cite{rombach2022stablediffusion}, we sample $Z$ using the reparametrization trick \cite{kingma2013auto} and regularize the latent space of the encoder $E_\theta$ using a Kullback--Leibler divergence term with a low coefficient and the total loss is thus:
\begin{equation}
\mathcal{L}(\theta, \phi) = \frac{1}{\vert \mathcal{I} \vert} \sum_{i=1}^{\vert \mathcal{I} \vert} \Big\Vert I_i - \mathcal{R} \Big( D_\phi\big(E_\theta(I_i, P_i)\big), C_i\Big) \Big\Vert_2^2 + \beta \, \mathcal{L}_{\text{KL}}(\theta).
\end{equation}

The use of a probabilistic framework accounts for the fact that multiple sets of Gaussians can yield the same projections, and provides a well behaved latent space.

\paragraph{Latent targets and conditioning.} We use \( E_\theta\) in two capacities:
given a dense set of input views, \(Z_{\text{target}} = E_\theta(\mathcal{I}_{\text{dense}})\) can provide targets for the 3D latent generative model. Given a single view or a small set of views \(\mathcal{I}_\text{sparse}\), \(Z_{\text{cond}} = E_\theta(\mathcal{I}_{\text{sparse}})\) provides model conditioning.
Thus, $E_\theta$ is trained to work with variable input set sizes.

\section{Sparse Auto-Regressive Modeling in 3D Latent Space}
\label{sec:painter_carver}
The encoder-decoder predicts 3D Gaussians covering only the observed parts of the scene. To generate geometry and scene content in unobserved parts, we introduce our sparse 3D latent generative model, see Fig.~\ref{fig:main2} for an overview.

\noindent \paragraph{Background: masked auto-regression.} 
Our generative model is built on Masked Auto-regressive models ~\cite{chang2022maskgit,li2022mage,li2024autoregressive}. 
Let $\mathbf{z} =(z_1, \hdots z_n)$, auto-regressive models decompose $P(\mathbf{z})$ using the chain rule of probabilities: $P(\mathbf{z}) = \prod_{i=1}^{n} P(z_i \mid z_{1}, \dots, z_{i-1})$. Masked auto-regressive models in particular can predict the chain in any order, and predict any group of variables from any other. They achieve this by training the model to take random visible sets of input and predicting the rest. To model the output distribution, a popular choice is a categorical distribution over discretized values; alternatively,\cite{li2024autoregressive} proposed a token-wise diffusion loss to model continuous values. 

\begin{figure*}[t]
    \centering
    \includegraphics[width=\linewidth]{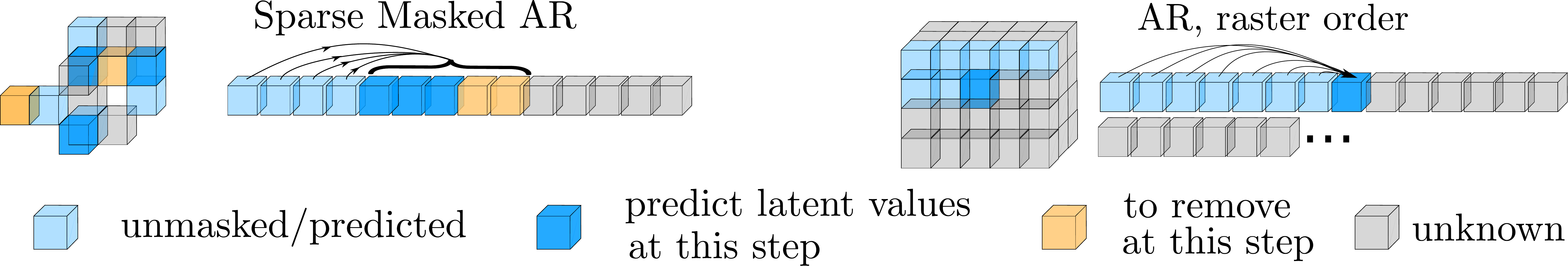}
    \caption{\textbf{Sparse 3D latent completion.} Sparse masked auto-regression is applied by jointly predicting latent token values as well as occupancies for a random observed subset of tokens. By contrast, raster-order auto-regressive models consider all tokens sequentially in the possible volume, which does not scale well.}
    \label{fig:train_gen}
\end{figure*}

\paragraph{Predicting occupancy and latent values.} 
Scene completion in a voxel space is a challenging task as sequences of unknown voxels can be of arbitrary length and voxels may be occupied or empty, making sequences spatially sparse. 
We thus propose a formulation where a masked auto-regressive model predicts both \emph{where} the 3D latent vectors through occupancy prediction, and their \emph{content} through latent prediction.
Concretely, our model \(H_{\psi} \) takes as input a sequence of conditioning voxel tokens from the encoder-decoder, a sequence of unknown voxel positions and predicts their occupancy as well as their latent values if they are occupied. It is composed of two bidirectional attention-based blocks $(H_\psi^1, H_\psi^2)$ and two lightweight voxel-wise heads: an occupancy head $h_{occ}$ and a denoising diffusion head $h_\epsilon$ \cite{li2024autoregressive}.
The first block aggregates information from conditioning and observed latent vectors while for the second block, masked voxel embeddings \( \mathbf{z}_{\text{m}} \) are added to the aggregated context tokens, and completed by $H_\psi^2$, inspired by the MAE architecture~\cite{he2022mae}.
Finally, occupancy $\mathbf{o}$ and latent vector $\mathbf{z}$ are predicted independently per masked voxel using the prediction heads.

\paragraph{Training procedure.} 
Training our generative model does not require any 3D ground-truth information and simply requires supervision in the form of \text{target} tokens generated from encoding dense views through our encoder.
To ease notations in what follows, we assume that these target tokens $\textbf{z}$ are obtained by encoding a dense set of multi-view images $\mathbf{z} = E_\theta(\mathcal{I}_\text{target})$. 
 An important point is that target sequences can be \emph{incomplete}. Indeed, using a dense set of multi-view inputs $\mathcal{I}_\text{target}$ does not \emph{guarantee} that the entire scene will be observed.
Thus at training, our \emph{target} token positions are separated into three categories: $\mathcal{F}$ the set of occupied positions in $\textbf{z}$, $\mathcal{E}$ the set of positions observed to be empty, and $\mathcal{U}$ the set of unobserved positions. Their disjoint union covers the full volume:
\begin{equation}   
\mathcal{V} = \mathcal{F} \sqcup \mathcal{E} \sqcup \mathcal{U}.
\end{equation}
In full generality, masked auto-regressive models are trained by selecting a mask, \ie a \emph{random} subset $\mathcal{O}$ of observed tokens \emph{from the target}, and predicting the unobserved \emph{target} positions $\mathcal{T} = \mathcal{V} \setminus \mathcal{O}.$  In our setting, we exploit the mask in one additional way: at training, we use the mask to \emph{hide} unobserved positions in $\mathcal{U}$, \ie we constrain positions in $\mathcal{U}$ to always be in $\mathcal{T}$. They are thus predicted by the network; however, they will be masked from the loss, as the target is unknown. In addition, our model also observes a sparse set of inputs through their encoded representation, denoted $\mathbf{z}^c= E_\theta(\mathcal{I}_{\text{cond.}})$; this observation is provided to the model by adding tokens in $\mathbf{z}^c$ to its inputs when their corresponding positions are not included in $\mathbf{z}_\mathcal{O}.$ See Fig.\ref{fig:train_gen} for an overview. Denoting $p_\mathcal{T}$ the target positions, a training forward pass  can be summarized as: 
\begin{equation}
\hat{\mathbf{o}}_{\mathcal{T}},\;
\hat{\mathbf{z}}_{\mathcal{T}}
=
H_\psi\!\left(\mathbf{z}_{\mathcal{O}},\mathbf{z}^c, p_\mathcal{T}\right
),
\end{equation}
 with $\mathcal{O} \cap \mathcal{T} = \varnothing$, $\mathcal{O} \cup \mathcal{T} \subset \mathcal{V}$, $\mathcal{U} \subset \mathcal{T}$, and $\mathcal{U}$ masked from the loss.

\noindent As cameras are randomly placed, we assume that the unknown set $\mathcal{U}$ is random and thus the model will learn to cover full latent representations on average. 

\begin{table*}[t]
\centering
\caption{\textbf{Novel view synthesis} evaluation with \textit{two} conditioning views (very wide baseline) at resolution $224\times224$ on the 3DFront and RealEstate10k datasets.}
\begin{tabular}{l|l|cccc|cccc}
\toprule
\multirow{2}{*}{Category} & \multirow{2}{*}{Method} & \multicolumn{4}{c|}{3DFront} & \multicolumn{4}{c}{RealEstate10k} \\
 &  & FID$\downarrow$ & PSNR$\uparrow$ & SSIM$\uparrow$ & LPIPS$\downarrow$ & FID$\downarrow$ & PSNR$\uparrow$ & SSIM$\uparrow$ & LPIPS$\downarrow$ \\
\midrule

\multirow{3}{*}{Reconstruction}
 & 3DGS \cite{kerbl20233dgaussiansplatting} & 260 & 8.65 & 0.14 & 0.79 & 271 & 7.92 & 0.12 & 0.79 \\
 & PixelSplat \cite{charatan2024pixelsplat} & 165 & 9.07 & 0.18 & 0.67 & 155 & 13.73 & 0.41 & 0.50 \\
 & DepthSplat \cite{xu2024depthsplat} & 110 & 13.76 & 0.49 & 0.55 & 82 & 13.25 & 0.46 & 0.41 \\

\midrule

\multirow{4}{*}{Generation}
 & DiffusioNeRF \cite{wynn-2023-diffusionerf} & 229 & 12.80 & 0.19 & 0.73 & - & - & - & - \\
 & MVSplat360 \cite{chen2024mvsplat360} & 111 & 9.72 & 0.35 & 0.70 & 66 & 15.40 & 0.49 & 0.40 \\

 & LatentSplat \cite{wewer24latentsplat} & 180 & 13.92 & 0.33 & 0.61 & 53 & 16.09 & 0.48 & 0.39 \\
& \textbf{\ours (ours)} & \textbf{59} & \textbf{15.18} & \textbf{0.62} & \textbf{0.50} & \textbf{41} & \textbf{16.72} & \textbf{0.57} & \textbf{0.36} \\
\bottomrule
\end{tabular}
\label{tab:NVS_2v}
\end{table*}

\paragraph{Inference.} 
At inference, positions are grouped into disjoint subsets 
$\mathcal{T}_0 \sqcup \dots \sqcup \mathcal{T}_n = \mathcal{V}$ 
and the model is run iteratively, where predictions on 
$\mathcal{T}_k$ are added to the conditioning set before predicting 
$\mathcal{T}_{k+1}$.
Given conditioning information $\mathbf{z}^c$ 
we select a subset $\mathcal{T}_1$ of masked target voxels;
$H_\psi$ outputs occupancy scores and latent predictions. If the occupancy score is above a threshold $\tau$, the latent prediction is added to the set of observed voxels; otherwise it is discarded. This 
is repeated iteratively by selecting a new subset $\mathcal{T}_i$ of masked target voxels at each iteration. Thus, the model refines the scene geometry while predicting scene content.

This construction can work with any random ordering. 
We opt for a region growing approach, where subsets are selected around the previous set until all positions are exhausted.
We perform mask selection (\ie autoregressive order)
by building a sparse symmetric k-NN graph with observed voxels $\textbf{z}^o$ taken as seeds, from which a breadth-first search (BFS) assigns a depth level $l(v)$ to each voxel (length of the shortest path from a seed to $v$). Candidate voxels are then partitioned into depth levels and the resulting joint latent probability is:
\begin{equation}
 P(\mathbf{z}) = \prod_{l=1}^{\max_v(l(v))} P\left(\mathbf{z}_{\{l(v)=l\}} | \mathbf{z}_{\{l(v)<l\}} \right). 
 \end{equation}
This ordering promotes spatial consistency at inference.

\paragraph{Training.}  
The occupancy head is optimized through a masked binary cross entropy loss between $\mathbf{o}_\text{target}$ and $\mathbf{o}_\text{pred}$:
$\mathcal{L}_{\text{occ}} = BCE(\hat{\mathbf{o}}_\mathcal{T}, \mathbf{o}_\mathcal{T}).$
Inspired by \cite{li2024autoregressive}, we parametrize the latent prediction head as a tokenwise denoising network and optimize it with a tokenwise diffusion loss, effectively learning to sample plausible unobserved tokens conditioned on observed tokens. 
It is based on a DDPM \cite{ho2020denoising} schedule with an epsilon formulation, leading to the following loss: $\mathcal{L}_\text{diff} = \mathbb{E}_{\epsilon \sim \mathcal{N}(0,I),t} \Vert \epsilon - h_\epsilon(Z^\text{mask}_\text{target}|f_t^2,t) \Vert_2^2,
$ where $\epsilon$ denotes the corruption noise and $t$ us the timestep in the noise schedule. We apply this loss on the set of unobserved target positions $\mathcal{T}$.
In parallel, we introduce a second diffusion head to refine conditioning tokens based on all decoded information $\textbf{z}^c$, it is optimized with the same $\mathcal{L}_\text{diff}$ loss but applied on conditioning tokens only. As this task is much more constrained than the main novel token prediction task, we apply a $stopgrad$ operation before the conditioning token refinement head. 
We keep the cost of this tokenwise diffusion part at minimum (a few token-wise MLP layers) to maintain scalability, but empirically observe an improvement compared to a fully auto-regressive formulation and keep it by default. 

\paragraph{Hierarchical occupancy prediction.}
To further improve efficiency, we propose to perform occupancy prediction in a hierarchical coarse-to-fine manner. In that case both stages share the masked auto-regressive overall architecture and principles defined earlier, but differ in the following way:
the coarse stage performs a dense coarse occupancy prediction from all the voxels in the scene while the fine stage takes as input this coarse set of occupied voxels and predicts a refined set of occupancies.
The coarse model is 
used with a \emph{high recall} occupancy threshold at inference while the fine model 
is used with a threshold chosen for a good precision-recall tradeoff, thus optimizing compute and accuracy.

\section{Experiments}
\label{sec:experiments} 

We first present training details and evaluation protocol (\cref{ssec:details}), which primarily consists in evaluating scene reconstruction from unconstrained cameras, and sparse sets of conditioning views, \eg $2$ views.
In this context where with 2 cameras randomly sampled, input views often have no overlap, typically making baselines struggle, we show the benefits of \ours through quantitative evaluations and qualitative examples (\cref{ssec:xp}).
Finally we provide a comprehensive set of ablations to justify our design choices (\cref{ssec:ablations}).

\begin{figure*}[t]
  \centering
  \includegraphics[width=1.0\linewidth]{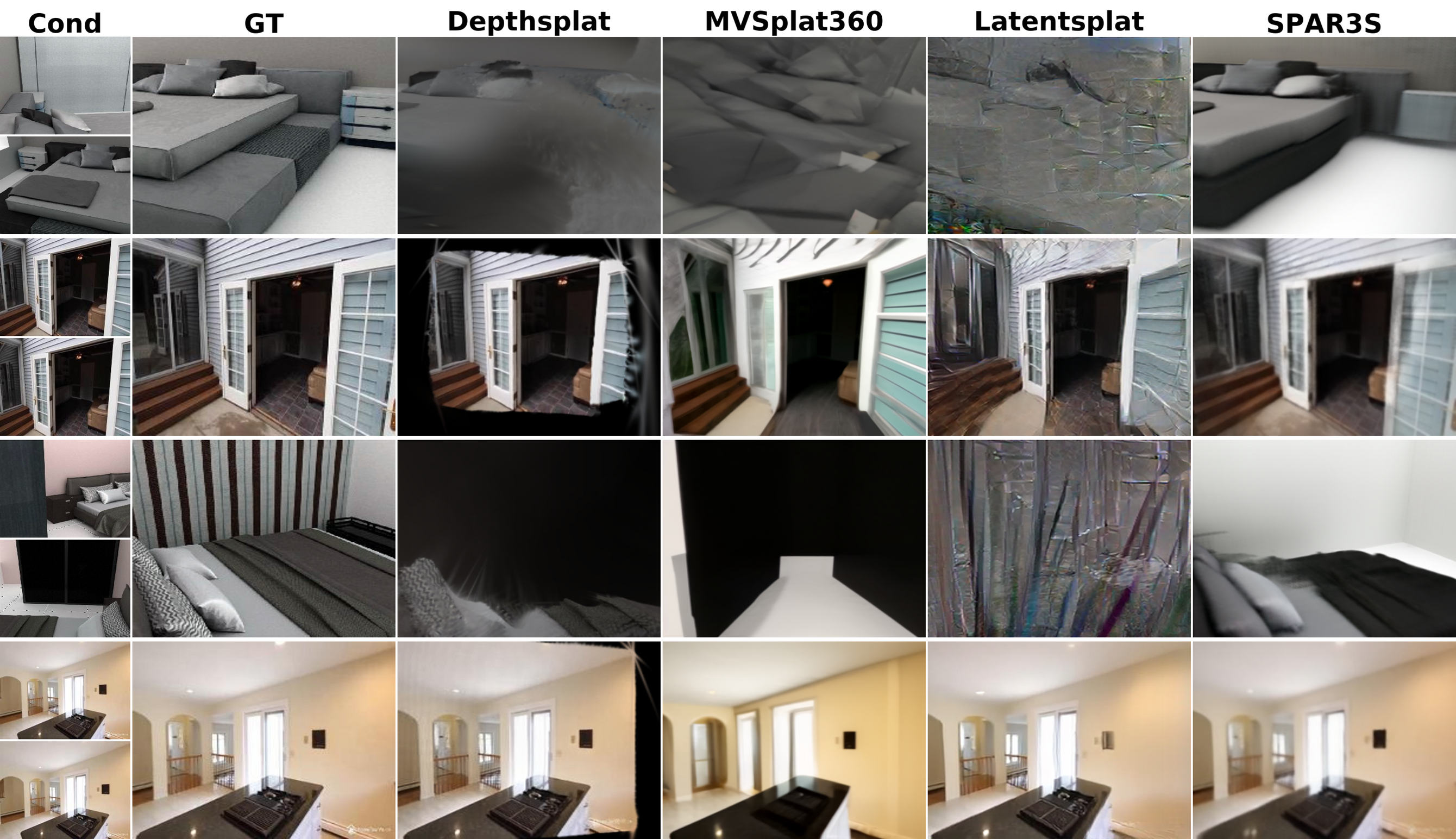}
  \caption{\textbf{Visual comparison} of novel rendered views from 2 conditioning images (left column). Overall, our approach handles conditioning views with little overlap and large novel view extrapolation (lines 1 and 3) while other 3D generative approaches  fail. On easier setups (lines 2 and 4), our model still provides better scene completions. }
  \label{fig:visu2V}
\end{figure*}

\subsection{Experimental details}
\label{ssec:details}

\paragraph{Datasets.}
We train and evaluate our models on a synthetic indoor dataset generated with 3DFront~\cite{fu2023dfront}, as well as the real-world RealEstate10K dataset~\cite{zhou2018stereo}.
Our synthetic dataset, referred to as 3DFront, comprises over $10 000$ indoor scenes covering bedrooms, living rooms, dining rooms, and libraries, with diverse layouts, furniture, and textures. For each scene, cameras are first distributed randomly. A subset of cameras that maximizes overall scene coverage is then selected as the final scene viewpoints. In the sparse-view setting, this naturally reduces the overlap between viewpoints, making novel-view synthesis significantly more challenging. We use ground-truth camera poses, and initial pointmaps are obtained by backprojecting pixels using rendered depth maps.

To assess our method’s ability to handle real data without ground-truth camera information, we also evaluate on RealEstate10K, which consists of over 80k real estate video clips, predominantly captured indoors. We estimate camera poses, intrinsics, and initial pointmaps using MASt3R-SfM~\cite{duisterhof2025mastrsfm}. Owing to their video origin, the camera paths in RealEstate10K follow smooth trajectories and produce densely sampled views with substantial overlap, making the novel-view synthesis task comparatively easier.

\paragraph{Training details.}
Our models are trained and evaluated at a resolution of $224$x$224$. 
Training is done from scratch on a single A100 GPU; training both the scene encoder-decoder and the latent prediction model takes approximately 4 days each in that setting. 
Given a training scene, we randomly sample $4$ conditioning views and 12 target views. Masking ratio in the target set of voxels is uniformly sampled in $[0.5,1]$. We refer to the supplementary material for further training details. 

\pgfplotstableread[row sep=\\,col sep=space]{%
cond idx c FID KID_mean KID_std ssim psnr lpips \\
1 1 1 47.058406 0.514451 0.051226 0.739900 15.412600 0.384100 \\
1 1 2 47.058406 0.514451 0.051226 0.739900 15.412600 0.384100 \\
1 1 3 46.013601 0.475891 0.057049 0.728900 15.080900 0.390500 \\
1 1 4 50.894944 0.461682 0.056241 0.723600 14.783700 0.394800 \\
1 1 5 52.705353 0.439175 0.057300 0.722400 14.634500 0.396000 \\
2 2 1 41.290115 0.563751 0.045105 0.778200 17.528200 0.341900 \\
2 2 2 41.290115 0.563751 0.045105 0.778200 17.528200 0.341900 \\
2 2 3 39.338271 0.547827 0.046140 0.762300 17.091900 0.355000 \\
2 2 4 40.360778 0.532553 0.052002 0.759000 16.868500 0.358500 \\
2 2 5 41.016414 0.537288 0.055900 0.758500 16.722400 0.358800 \\
4 3 1 39.90134 0.612031 0.037269 0.795400 18.654000 0.320000 \\
4 3 2 37.843754 0.571985 0.038295 0.787700 18.327600 0.326700 \\
4 3 3 38.483646 0.581555 0.040840 0.782400 18.240600 0.333300 \\
4 3 4 39.199613 0.567090 0.045474 0.778100 17.999700 0.336500 \\
4 3 5 38.850151 0.556051 0.045573 0.777300 17.991600 0.336700 \\
8 4 1 38.731737 0.609759 0.036004 0.791800 18.796800 0.316800 \\
8 4 2 38.731737 0.582747 0.042978 0.790600 18.736600 0.321400 \\
8 4 3 39.257222 0.589211 0.044378 0.790100 18.754500 0.323800 \\
8 4 4 39.065724 0.588211 0.041048 0.788500 18.675300 0.325400 \\
8 4 5 40.144588 0.588153 0.045509 0.787200 18.574300 0.328200 \\
12 5 1 41.899147 0.604714 0.040707 0.796400 18.956700 0.309200 \\
12 5 2 41.240050 0.594004 0.040707 0.797500 19.073000 0.318200 \\
12 5 3 40.033710 0.592005 0.036128 0.791300 18.832200 0.322600 \\
12 5 4 40.932346 0.591390 0.034760 0.791800 18.876500 0.322700 \\
12 5 5 40.909179 0.592357 0.039681 0.790500 18.824800 0.324200 \\
}\datatable

\pgfplotstableread[row sep=\\,col sep=space]{%
cond_f idx_f bce_f FID_f KID_mean_f KID_std_f ssim_f psnr_f lpips_f \\
1 1 1 120.14 0.14 nan 0.7172 11.73 0.5977 \\
1 1 2 87.24 0.11 nan 0.7111 9.83 0.5990 \\
1 1 3 101.09 0.10 nan 0.7129 9.46 0.5955 \\
1 1 4 113.22 0.11 nan 0.7159 9.08 0.5928 \\
1 1 5 87.40 0.14 nan 0.7409 12.74 0.5635 \\
2 2 1 68.26 0.13 nan 0.7391 12.11 0.5664 \\
2 2 2 77.71 0.11 nan 0.7420 11.65 0.5589 \\
2 2 3 86.84 0.11 nan 0.7466 11.39 0.5538 \\
2 2 4 96.01 0.11 nan 0.7475 11.01 0.5534 \\
2 2 5 59 0.18 nan 0.79 15.24 0.5001 \\
4 3 1 58.23 0.17 nan 0.7807 14.55 0.5157 \\
4 3 2 53.03 0.15 nan 0.7828 14.27 0.5108 \\
4 3 3 69.21 0.14 nan 0.7848 13.87 0.5067 \\
4 3 4 73.89 0.13 nan 0.7858 13.64 0.5061 \\
4 3 5 52 0.23 nan 0.83 18.00 0.46 \\
8 4 1 51 0.19 nan 0.8272 17.75 0.4565 \\
8 4 2 52.66 0.18 nan 0.8276 17.54 0.4534 \\
8 4 3 54.26 0.17 nan 0.8283 17.22 0.4520 \\
8 4 4 57.04 0.16 nan 0.8285 16.92 0.4520 \\
8 4 5 56.51 0.27 nan 0.85 19.78 0.43 \\
12 5 1 51.52 0.23 nan 0.8502 19.62 0.4262 \\
12 5 2 50.97 0.21 nan 0.8515 19.42 0.4230 \\
12 5 3 50.94 0.20 nan 0.8526 19.28 0.4219 \\
12 5 4 51.65 0.20 nan 0.8533 19.12 0.4212 \\
12 5 5 51.65 0.20 nan 0.8533 19.12 0.4212 \\
}\datatable

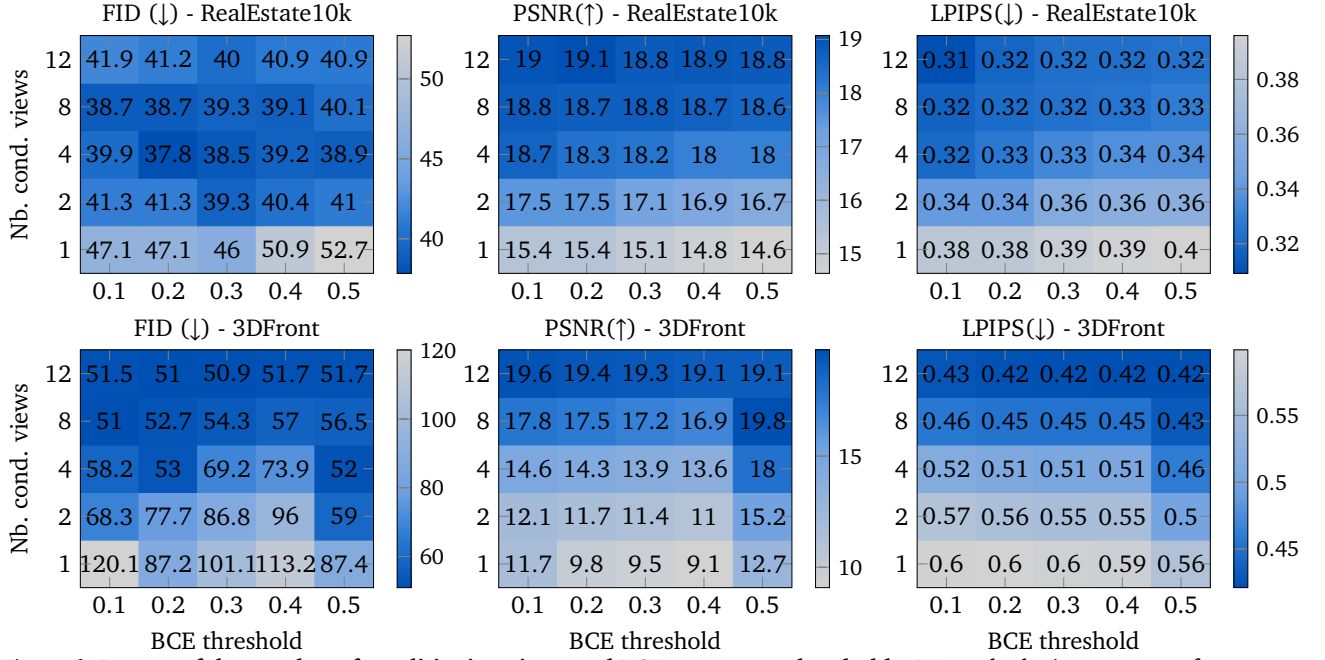
\begin{figure*}
\resizebox{\linewidth}{!}{%
\begin{tikzpicture}[baseline]
\begin{groupplot}[
    height=5cm,
    group style={
        group size=3 by 2,
        horizontal sep=1.8cm,
        vertical sep=1.1cm,
    },
    view={0}{90},
    xlabel={BCE threshold},
    ylabel={Nb. cond. views},
    colorbar,
    colorbar right=0mm,
    colorbar style={
        width=2mm,
        yticklabel style={font=\small},
    },
    colormap={blueorange}{
    rgb255(0.0cm) = (  0, 80,180)   
    rgb255(0.2cm) = ( 40,120,210)   
    rgb255(0.4cm) = (110,160,225)   
    rgb255(1.0cm) = (210,210,210)   
  },
    enlargelimits=false,
    xmin=0.5, xmax=5.5,
    ymin=0.5, ymax=5.5,
    ytick={1,2,3,4,5},
    yticklabels={1,2,4,8,12},
    xtick={1,2,3,4,5},
    xticklabels={0.1,0.2,0.3,0.4,0.5},
    unbounded coords=discard,
    axis on top,
]

\nextgroupplot[title={FID ($\downarrow$) - RealEstate10k}, title style={yshift=-0.2cm}, xlabel={}]
\addplot[
    matrix plot*,
    point meta=explicit,
    nodes near coords,
    nodes near coords align={center},
    nodes near coords style={font=\normalsize},
    nodes near coords={\pgfmathfloatifflags{\pgfplotspointmeta}{3}{}{\pgfmathprintnumber[fixed,precision=1]{\pgfplotspointmeta}}},
] table[
    x=c,
    y=idx,
    meta=FID,
] {\datatable};

\nextgroupplot[title={PSNR($\uparrow$) - RealEstate10k},ylabel={}, title style={yshift=-0.2cm}, xlabel={}, colormap={blueorange_rev}{
    rgb255(0.0cm) = (210,210,210)
    rgb255(0.6cm) = (110,160,225)
    rgb255(0.8cm) = ( 40,120,210)
    rgb255(1cm)   = (  0, 80,180)
  },]
\addplot[
    matrix plot*,
    point meta=explicit,
    nodes near coords,
    nodes near coords align={center},
    nodes near coords style={font=\normalsize},
    nodes near coords={\pgfmathfloatifflags{\pgfplotspointmeta}{3}{}{\pgfmathprintnumber[fixed,precision=1]{\pgfplotspointmeta}}},
] table[
    x=c,
    y=idx,
    meta=psnr,
] {\datatable};

\nextgroupplot[title={LPIPS($\downarrow$) - RealEstate10k},ylabel={}, title style={yshift=-0.2cm},xlabel={}]
\addplot[
    matrix plot*,
    point meta=explicit,
    nodes near coords,
    nodes near coords align={center},
    nodes near coords style={font=\normalsize},
    nodes near coords={\pgfmathfloatifflags{\pgfplotspointmeta}{3}{}{\pgfmathprintnumber[fixed,precision=2]{\pgfplotspointmeta}}},
] table[
    x=c,
    y=idx,
    meta=lpips,
] {\datatable};

\nextgroupplot[title={FID ($\downarrow$) - 3DFront}, title style={yshift=-0.2cm}, ]
\addplot[
    matrix plot*,
    point meta=explicit,
    nodes near coords,
    nodes near coords align={center},
    nodes near coords style={font=\normalsize},
    nodes near coords={\pgfmathfloatifflags{\pgfplotspointmeta}{3}{}{\pgfmathprintnumber[fixed,precision=1]{\pgfplotspointmeta}}},
] table[
    x=bce_f,
    y=idx_f,
    meta=FID_f,
] {\datatable};

\nextgroupplot[title={PSNR($\uparrow$) - 3DFront},ylabel={}, title style={yshift=-0.2cm}, colormap={blueorange_rev}{
    rgb255(0.cm) = (210,210,210)
    rgb255(0.6cm) = (110,160,225)
    rgb255(0.8cm) = ( 40,120,210)
    rgb255(1cm)   = (  0, 80,180)
  },]
\addplot[
    matrix plot*,
    point meta=explicit,
    nodes near coords,
    nodes near coords align={center},
    nodes near coords style={font=\normalsize},
    nodes near coords={\pgfmathfloatifflags{\pgfplotspointmeta}{3}{}{\pgfmathprintnumber[fixed,precision=1]{\pgfplotspointmeta}}},
] table[
    x=bce_f,
    y=idx_f,
    meta=psnr_f,
] {\datatable};

\nextgroupplot[title={LPIPS($\downarrow$) - 3DFront}, title style={yshift=-0.2cm}, ylabel={}]
\addplot[
    matrix plot*,
    point meta=explicit,
    nodes near coords,
    nodes near coords align={center},
    nodes near coords style={font=\normalsize},
    nodes near coords={\pgfmathfloatifflags{\pgfplotspointmeta}{3}{}{\pgfmathprintnumber[fixed,precision=2]{\pgfplotspointmeta}}},
] table[
    x=bce_f,
    y=idx_f,
    meta=lpips_f,
]{\datatable};
\end{groupplot}
\end{tikzpicture}}  \\[-0.3cm]
\caption{\textbf{Impact of the number of conditioning views and BCE occupancy thresholds.} We study the impact on performance of varying the threshold used for binary occupancy prediction (in $\{0.1, 0.2, 0.3, 0.4, 0.5\})$ for various numbers of conditioning views ($\{1, 2, 4, 8, 12\}$). Increasing the number of conditioning views consistently improves performance, while lower threshold tend to give gains. With PSNR and LPIPS we see a diagonal pattern: with more conditioning views, the threshold can be increased as geometry prediction gets more accurate.
}
    \label{fig:cond_bce_plot}
\end{figure*}

\paragraph{Evaluation protocol.}
We evaluate the ability to reconstruct scene geometry and appearance by performing novel view synthesis from a sparse set of observed images. We report standard image-level reconstruction metrics—PSNR and SSIM—along with the perceptual metric LPIPS~\cite{zhang2018perceptual}, all computed on rendered novel views. To further assess the visual quality and diversity, we measure FID~\cite{heusel2017gans} and KID~\cite{binkowski2018demystifying} across datasets.
The number of conditioning views varies among ${2,4,8,12,16}$ depending on the experiment, and we explicitly indicate the chosen setting for each evaluation. 
Given a scene, conditioning views are first randomly sampled, novel views are then randomly sampled among the remaining images. We deliberately adopt this random sampling strategy for both our evaluation datasets such that the splits span a wide range of difficulty, from easy novel views containing overlap with the conditioning set to challenging novel views that share little or no overlap (as opposed to `easier' splits that may be found in the literature \cite{charatan2024pixelsplat} that contain substantial overlaps).

\paragraph{Baselines.}
We primarily compare our approach against recent state-of-the-art feed-forward generative 3D Gaussian scene reconstruction methods: Latent\-Splat~\cite{wewer24latentsplat} which employs a variational latent representation within a VAE framework, enabling more reliable reconstruction outside the observed camera frustums as well as MVSplat360~\cite{chen2024mvsplat360} which samples images with a video diffusion model operating in a latent space spanned by rendering 3DGS latent vectors predicted from a feed-forward 3DGS model. We further include the 3D generative DiffusioNeRF~\cite{wynn-2023-diffusionerf} which leverages generative priors from diffusion models when reconstructing scenes with neural radiance fields. 
For the sake of completeness, we include the following pixel-aligned methods, PixelSplat~\cite{charatan2024pixelsplat}, a well-established model that regresses 3D Gaussians along viewing rays from paired context images as well as Depthsplat~\cite{xu2024depthsplat} which leverages geometric priors resulting in more accurate reconstruction. Finally, vanilla 3DGS~\cite{kerbl20233dgaussiansplatting} is also included as a reference.

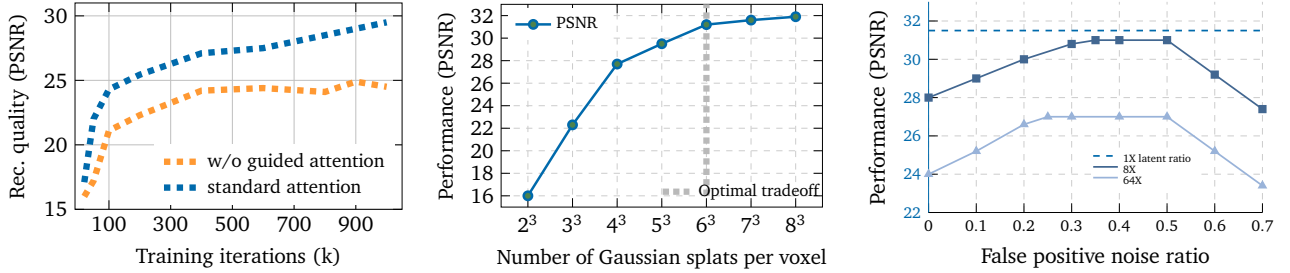
\begin{figure*}[t]
    \centering
    
    \resizebox{!}{103pt}{\begin{tikzpicture}
\begin{axis}[
    width=7cm, height=5cm,
    xlabel={Training iterations (k)},
    ylabel={Rec. quality (PSNR)},
    xmin=-10, xmax=1050,
    ymin=15, ymax=31,
    xtick={100,300,500,700,900},
    ytick={15, 20, 25, 30},
    grid=both,
    grid style={gray!30},
    major grid style={line width=.2pt, gray!50},
    tick style={black},
    line width=1pt,
    legend style={
        at={(0.98,0.02)},
        anchor=south east,
        cells={anchor=west},
        draw=none,
        font=\small
    },
    every axis plot/.append style={thick},
]

\addplot[
    color=orange!80!white,
    dashed,
    line width=3pt
] coordinates {
    (20, 16)
    (50,17.2)
    (100,21.1)
    (200, 22.3)
    (400, 24.2)
    (600, 24.4)
    (800, 24.1)
    (900, 24.9)
    (1000, 24.5)
};

\addplot[
    color=cyan!45!blue!90!black,
    dashed,
    line width=3pt
] coordinates {
    (20,17.1)
    (50,22.)
    (100, 24.3)
    (200, 25.44)
    (400, 27.1)
    (600, 27.5)
    (800, 28.5)
    (900, 29.0)
    (1000, 29.5)
};

\addlegendentry{w/o guided attention}
\addlegendentry{standard attention}

\end{axis}
\end{tikzpicture}}\hfill
    \resizebox{!}{103pt}{\begin{tikzpicture}
\begin{axis}[
    width=7cm, height=5cm,
    xlabel={Number of Gaussian splats per voxel},
    ylabel={Performance (PSNR)},
    symbolic x coords={$2^{3}$,$3^{3}$,$4^{3}$,$5^{3}$,$6^{3}$,$7^{3}$,$8^{3}$},
    xtick=data,
    ymin=15, ymax=33,
    ytick={16,18,20,22,24,26,28,30,32},
    grid=both,
    grid style={dashed, gray!25},
    major grid style={line width=.2pt, gray!40},
    line width=1pt,
    ylabel style={text=black},
    xlabel style={text=black},
    yticklabel style={text=black},
    tick style={black},
    every axis plot/.append style={thick},
    legend style={
        at={(0.02,0.98)}, anchor=north west,
        draw=none, font=\footnotesize, row sep=-2pt
    },
]

\addplot[
    color=cyan!45!blue!90!black,
    mark=*,
    mark options={fill=green!30!black!70!white},
    line width=1.2pt
] coordinates {
    ($2^{3}$,16.0)
    ($3^{3}$,22.3)
    ($4^{3}$,27.7)
    ($5^{3}$,29.5)
    ($6^{3}$,31.2)
    ($7^{3}$,31.6)
    ($8^{3}$,31.9)
};
\addlegendentry{PSNR}

\addplot[
    densely dashed,
    line width=3pt,
    gray!55
] coordinates {($6^{3}$,16) ($6^{3}$,33)};

\end{axis}

\node[anchor=south east, font=\footnotesize, xshift=-1pt, yshift=30pt]
  at (current bounding box.south east) {%
  \begin{tikzpicture}
    \draw[densely dashed, line width=3pt, gray!55] (0,0) -- (0.45,0);
    \node[anchor=west] at (0.5,0) {Optimal tradeoff};
  \end{tikzpicture}
};

\end{tikzpicture}}\hfill
    \resizebox{!}{103pt}{\begin{tikzpicture}

\begin{axis}[
    name=combo,
    width=7cm, height=5cm,
    xlabel={False positive noise ratio},
    xlabel style={text=black, font=\scriptsize}, 
    ylabel={Performance (PSNR)},
    ylabel style={
        text=black, 
        font=\fontsize{6.5pt}{8pt}\selectfont,
        yshift=-0.5cm,
    },
    ticklabel style={font=\scriptsize},            
    yticklabel style={text=cyan!45!blue!90!black},
    axis lines=box,                            
    xmin=0, xmax=0.7,    
    xtick={0, 0.1, 0.2, 0.3, 0.4, 0.5, 0.6, 0.7},
    ymin=22, ymax=33,    
    ytick={22,24,26,28,30,32},
    grid=both,
    grid style={dashed, gray!25},
    major grid style={line width=.2pt, gray!40},
    line width=1pt,
    axis y line*=left,
    axis x line*=bottom,
    y axis line style={cyan!45!blue!90!black},
    tick style={black},
    every axis plot/.append style={thick},
    legend cell align={left},
    legend style={
        at={(0.80,0.2)},        
        anchor=east,            
        draw=none,
        font=\tiny,                                
        row sep=-4pt,
        inner sep=0pt,
        legend image code/.code={
            \draw[#1] (0cm,0.06cm) -- (0.4cm,0.06cm);
        }
    },
]

\addplot[color=cyan!45!blue!90!black, dashed, no markers] coordinates {
    (0.0, 31.5) (0.7, 31.5)
};
\addlegendentry{1X latent ratio}

\addplot[color=cyan!45!blue!90!black!70!black, mark=square*, mark size=1.5pt] coordinates {
    (0.0, 28.0) (0.1, 29.0) (0.2, 30.0) (0.3, 30.8) (0.35, 31.0) 
    (0.4, 31.0) (0.5, 31.0) (0.6, 29.2) (0.7, 27.4)
};
\addlegendentry{8X}

\addplot[color=cyan!45!blue!90!black!40, mark=triangle*, mark size=1.8pt] coordinates {
    (0.0, 24.0) (0.1, 25.2) (0.2, 26.6) (0.25, 27.0) (0.3, 27.0) 
    (0.4, 27.0) (0.5, 27.0) (0.6, 25.2) (0.7, 23.4)
};
\addlegendentry{64X}

\end{axis}
\end{tikzpicture}}
    \caption{
    \textbf{Ablation studies.}
    \textbf{Left:} Impact of guided attention on training dynamics — adding attention logits is beneficial.
    \textbf{Middle:} Impact of the number of splats per voxel — diminishing returns are observed; we use $6^3$ splats per voxel.
    \textbf{Right:} Latent compression ratio and false positives in learned upsampling: approximately 25\% of false positives is optimal.
    }
    \label{fig:ablations_combined}
\end{figure*}

\subsection{Two-view NVS}
We first evaluate scene reconstruction from two conditioning views, a challenging setting in which most of the scene is unobserved and must be inferred from minimal context. 
Novel-view synthesis results on 3DFront and RealEstate10K are reported in Tab.~\ref{tab:NVS_2v}.
As expected, both optimization-based~\cite{kerbl20233dgaussiansplatting} and feed-forward 3DGS~\cite{charatan2024pixelsplat}  methods fail to produce meaningful reconstructions, due to their limited ability to extrapolate beyond the frustums of conditioning views, and their difficulty in reproducing observed regions from novel views with significant viewpoint variation. Depthsplat \cite{xu2024depthsplat}, with geometric priors, is able to better handle large viewpoint changes but still inherently cannot reconstruct unobserved areas.
LatentSplat~\cite{wewer24latentsplat} benefits from its generative formulation and shows improved reconstructions in novel areas. However, it remains constrained by its epipolar-based encoder architecture which prevents it from handling overly wide viewpoint variations (as illustrated by the clear performance drop between Re10k and 3DFront). Similarly, MVSplat360 diffusion model~\cite{chen2024mvsplat360} struggles to handle unconstrained cameras poses which leads to inconsistent reconstructions.
In contrast, the combination of an autoregressive generative formulation and a sparse voxelized latent space allows SPAR3S to substantially outperform all 3D generative baselines on this challenging wide-two views scene reconstruction. Qualitative comparisons are shown in Fig.~\ref{fig:visu2V}. 

\begin{figure*}[t]
\centering
\begin{minipage}[t]{0.48\linewidth}
    \centering
    \vspace*{0.1cm}
    \resizebox{\linewidth}{!}{\begin{tikzpicture}
\begin{axis}[
    width=7.3cm, height=6cm,
    view={0}{90},
    axis on top,
    enlargelimits=false,
    xlabel={Latent channel dimension},
    ylabel={KL coefficient $\beta$},
    xmin=0.5, xmax=4.5,
    ymin=0.5, ymax=4.5,
    xtick=data,
    ytick=data,
    xticklabels={16,32,64,128},
    yticklabels={$10^{-3}$,$10^{-2}$,$10^{-1}$,$1$},
    ticklabel style={font=\small},
    colormap={blueorange}{
        rgb255(0cm)=(255,165,115);
        rgb255(1cm)=(80,185,255);
    },
    colorbar,
    grid=none
]
\addplot[
    matrix plot*,
    mesh/cols=4,
    point meta=explicit,
    draw=black,
    nodes near coords={\pgfmathprintnumber[fixed,precision=2]\pgfplotspointmeta},
    every node near coord/.append style={font=\footnotesize, text=black, anchor=center},
] table [meta=val] {
x  y  val
1  1  28.1
2  1  31.2
3  1  31.3
4  1  31.4
1  2  27.5
2  2  31.3
3  2  31.4
4  2  31.5

1  3  27.4
2  3  31.2
3  3  31.3
4  3  31.4

1  4  24.2
2  4  25.6
3  4  25.5
4  4  26.1
};
\draw[black, very thick]
  ([yshift=-5pt]axis cs:1.8,3) -- ([yshift=-5pt]axis cs:2.2,3);
\end{axis}
\end{tikzpicture}}
    \caption{\textbf{Information bottlenecks on the latent variable.} Impact measured with PSNR.  
    A clear trend emerges: a KL coefficient stronger than $0.1$ or a latent 
    dimension below $32$ both significantly degrade performance; otherwise 
    reconstruction accuracy is stable over a wide range. }
    \label{fig:bottleneck_plot}
\end{minipage}
\hfill
\begin{minipage}[t]{0.48\linewidth}
    \centering
    \captionof{table}{\textbf{Ablation study} of the autoregressive diffusion model 
    on the 3DFRONT dataset with four conditioning views. Each row removes 
    one component of the full model. Metrics are reported on novel views.}
    \resizebox{\linewidth}{!}{
    \begin{tabular}{lccc}
    \toprule
    Configuration 
    & PSNR$\uparrow$ 
    & SSIM$\uparrow$ 
    & LPIPS$\downarrow$ \\
    \midrule
    \textbf{\ours (full model)} & \textbf{15.18} & \textbf{0.62} & \textbf{0.50} \\
    \midrule
    \multicolumn{4}{l}{Ablations (each removes one component)} \\
    ~~~~w/o diff.           & 13.90 & 0.53 & 0.54 \\
~~~~w/o 3D RPE          & 13.74 & 0.49 & 0.58 \\
~~~~w/o BFS ordering    & 14.81 & 0.59 & 0.53 \\
~~~~w/o occ. reference  & 14.48 & 0.56 & 0.56 \\
\midrule
\multicolumn{4}{l}{Additional Oracle Variant} \\
~~~~\textcolor{gray}{\ours w/ gt. occ.} 
& \textcolor{gray}{17.90} 
& \textcolor{gray}{0.68} 
& \textcolor{gray}{0.45} \\
    \bottomrule
    \end{tabular}
    }
    
    \label{tab:abl_painter_carver}
\end{minipage}

\end{figure*}

\subsection{Multi-view NVS}
\label{ssec:xp}

In Fig.~\ref{fig:cond_bce_plot} we increase the amount of contextual information the model receives by increasing the number of conditioning views provided to the models, using $n=\{1, 2, 4, 8, 12\}$. The reconstruction task thus becomes more constrained, inducing easier scene completion in unobserved areas. This is indeed illustrated by the improving NVS metrics as the number of conditioning views increases. We also vary the threshold used in binary occupancy prediction. Overall, performance tends to be better with low thresholds: missing out parts of the room degrades metrics more than adding spurious parts. We notice a diagonal pattern: the model makes more accurate occupancy predictions with more views.

\subsection{Ablations}
\label{ssec:ablations}
\paragraph{Encoder-Decoder.}
We study the main design choices of the scene encoder-decoder. 
First, in Fig.~\ref{fig:ablations_combined} (left) we show the impact of removing the bincount-based guided attention mechanism, which results in slower training and lower performance. Second, 
we show in the middle plot the impact of the number of Gaussian regressed on PSNR. While the performance improves monotonically, there is a clear diminishing return and thus we chose a value of $6^3$. 
Third, we show in the right plot that when spatial compression is used, injecting false positives at training is key to obtain optimal performance. Around 25\% is a robust value for both $8\times$ and $64\times$ compression ratios.
Finally, we evaluate different information bottlenecks on the encoded latent variables in Fig.~\ref{fig:bottleneck_plot}; KL coefficients are taken in $ \beta \in \{0.001, 0.01, 0.1, 1\}$ and latent dimension in $\{16, 32, 64, 128\}.$  A KL coefficient stronger than $0.1$ or a latent
dimension below $32$ both significantly degrade performance;  we pick the tighter possible bottleneck: $\beta = 0.1$ and $\text{dim}(z)=32.$

\paragraph{Autoregressive diffusion model.} We ablate different components of the autoregressive diffusion model and report novel view rendering results in Tab.~\ref{tab:abl_painter_carver}. Decoding latent vectors with a linear head (\emph{wo. diff}) instead of a diffusion head degrades the model's ability to sample coherent tokens in unobserved areas. Similarly, injecting relative positional encoding in self-attention layers proves important for spatial reasoning when decoding the scene (\emph{w/o. 3D RPE}). During inference, using a random autoregressive order (\emph{w/o. BFS ordering}) and not performing occupancy refinement (\emph{w/o. occ. reference}) also degrades the consistency and quality of the predicted Gaussians. 
Finally, in order to evaluate the model's ability to reconstruct latent tokens without the influence of occupancy predictions, we decode the scenes using ground truth occupancies (\emph{\ours w. gt. occ.}). Unsurprisingly, this improves the quality of the results.

\section{Conclusion}
\label{sec:conclusions}

We introduce \ours, a novel generative model for 3D scene reconstruction. 
\ours processes sparse and unconstrained input views and encodes them into a 3D latent space associated with a voxel-aligned 3DGS encoder-decoder. 
\ours further performs full scene completion in this 3D latent space by leveraging an efficient autoregressive architecture that jointly infers geometry and latent content to exploit scene sparsity. Our results demonstrate the potential of autoregressive latent modeling for 3D scene generation. As such, our work opens promising avenues for applying masked autoregressive models to 3D data and for advancing generative reconstruction in sparse-view scenarios.

\clearpage
{
    \small
    \bibliographystyle{ieeenat_fullname}
    \bibliography{main}
}

\clearpage
\appendix

In this supplementary material, we provide additional implementation and training details in Sec.~\ref{sec:more_details}. We also 
display more visualizations in Sec.~\ref{sec:supp_visus}. In addition, we discuss on limitations of our method in Sec.~\ref{sec:limitation}.

\section{Additional implementation details}
\label{sec:more_details}
\paragraph{Architecture of the 3D Autoregressive model.}
 \(H_{\psi} \) is composed of two bidirectional attention-based blocks $\{ H_{\psi}^1, H_{\psi}^2\}$ and two lightweight voxel-wise heads: an occupancy head $h_{occ}$ and a denoising diffusion head $h_\epsilon$~\cite{li2024autoregressive}.
 Both $H_{\psi}^1$ and $H_{\psi}^2$ contain 5 self-attention blocks which include a LayerNorm~\cite{ba2016layer}, an MLP (Multi-Layer Perception) and a multi-head self-attention layer. 
 The input sequence to $H_{\psi}^1$ is composed of conditioning tokens and unmasked tokens which are projected to $H_{\psi}^1$'s dimension with separate linear layers and LayerNorm. Similarly, the input sequence to $H_{\psi}^2$ is composed of the output tokens of $H_{\psi}^1$ and masked tokens, both are projected to have the same dimension as $H_{\psi}^2$ with separate linear layers and LayerNorm.
 Absolute positional encoding is added to any token within these input sequences by applying sinusoidal encoding on the normalized voxel coordinates. 
 Furthermore, 3D
 RoPE~\cite{su2021roformer} using voxel coordinates, is applied in the self-attention layers. 
Masked tokens are initialized with a learnable \emph{mask} token. The internal dimension of $\{ H_{\psi}^1, H_{\psi}^2\}$ is set to 512. 

\paragraph{Implementation of the prediction heads.} 
The diffusion head $h_\epsilon$ follows the architecture in \cite{li2024autoregressive} with 2 layers of dimension 512. Conditioning information is injected through ada-ln modulation.
We apply signal to noise ratio (SNR) weighting in the diffusion objective to avoid over-weighting high noise timesteps and improve sample at lower noise timesteps. The occupancy head $h_{occ}$ is an MLP composed of three linear layers and GELU activations which outputs a single logit. It is optimized with a focal loss \cite{lin2017focal} to help mitigate the imbalance between occupied and empty voxels.

\paragraph{Occupancy prediction and coarse-to-fine refinement.} Predicting occupancy at the voxel level becomes computationally expensive because the number of voxels grows cubically with scene size. When training the encoder-decoder or the latent regression model, we may use "ground truth" occupancies deduced from the inputs to directly sparsify sequences. However, when training the occupancy model, this is obviously not possible anymore and we need to operate on the dense volume.
To address the cost of such dense 3D computations, we propose several optimizations. 
First, we predict occupancies in the compact latent space which strongly limits the scene size. We directly downsample ground truth to the latent space resolution to supervise the model.
Second, we adopt a coarse-to-fine strategy for occupancy prediction. 
Both the coarse and fine stages follow our 3D autoregressive latent framework for both training and inference (see \cref{sec:painter_carver} of the main paper).
The coarse stage is lightweight (\eg, two layers instead of five), and performs occupancy prediction over all voxels in the scene. During inference, it uses a high-recall threshold to ensure that potentially occupied regions are not missed. Latent prediction is also deactivated in the coarse stage to further improve efficiency.
The fine stage then operates only on the voxels selected by the coarse stage. The full model refines these initial occupancy predictions while predicting latent vectors when voxels are classified as occupied. It uses a threshold tuned for a strong precision-recall balance. This two-stage process significantly reduces computation while improving final accuracy. 
For further efficiency, we apply a MAR style random mask to \emph{both} inputs and targets for the MAR occupancy prediction: the targets always include the inputs, but do not cover the full tensor. 
We use a random masking ratio sampled between $0.3$ and $1$. We find this has no noticeable impact on final performance, while substantially reducing the memory computational cost.

\paragraph{Training Details.}
Training \ours involves two sequential steps. 
We first train the scene encoder for 1.8M iterations with batch size $4$, which takes approximately one week on a single NVIDIA A100 GPU. We use a cosine learning rate decay starting from $10^{-5}$ and the Adam optimizer~\cite{kingma2015adam}. The model is trained from scratch after a Xavier initialization~\cite{glorot2010understanding}.
We then train the autoregressive latent prediction model for 1.6M iterations with a batch size of 4 on a single NVIDIA V100 GPU. A cosine learning rate decay starting from $10^{-4}$ and the Adam optimizer~\cite{kingma2015adam} are used. The trained scene encoder provides conditioning tokens as well as target tokens for the latent prediction model.
The 3D latent space dimension is set to 32. In the encoder, after sampling the latents, we apply a group normalization with 8 groups. This regularizes the latent space and allows for more stable latent diffusion. 

\section{Visualizations} 
\label{sec:supp_visus}
\begin{figure*}[t]
  \centering
  \includegraphics[width=1.0\linewidth]{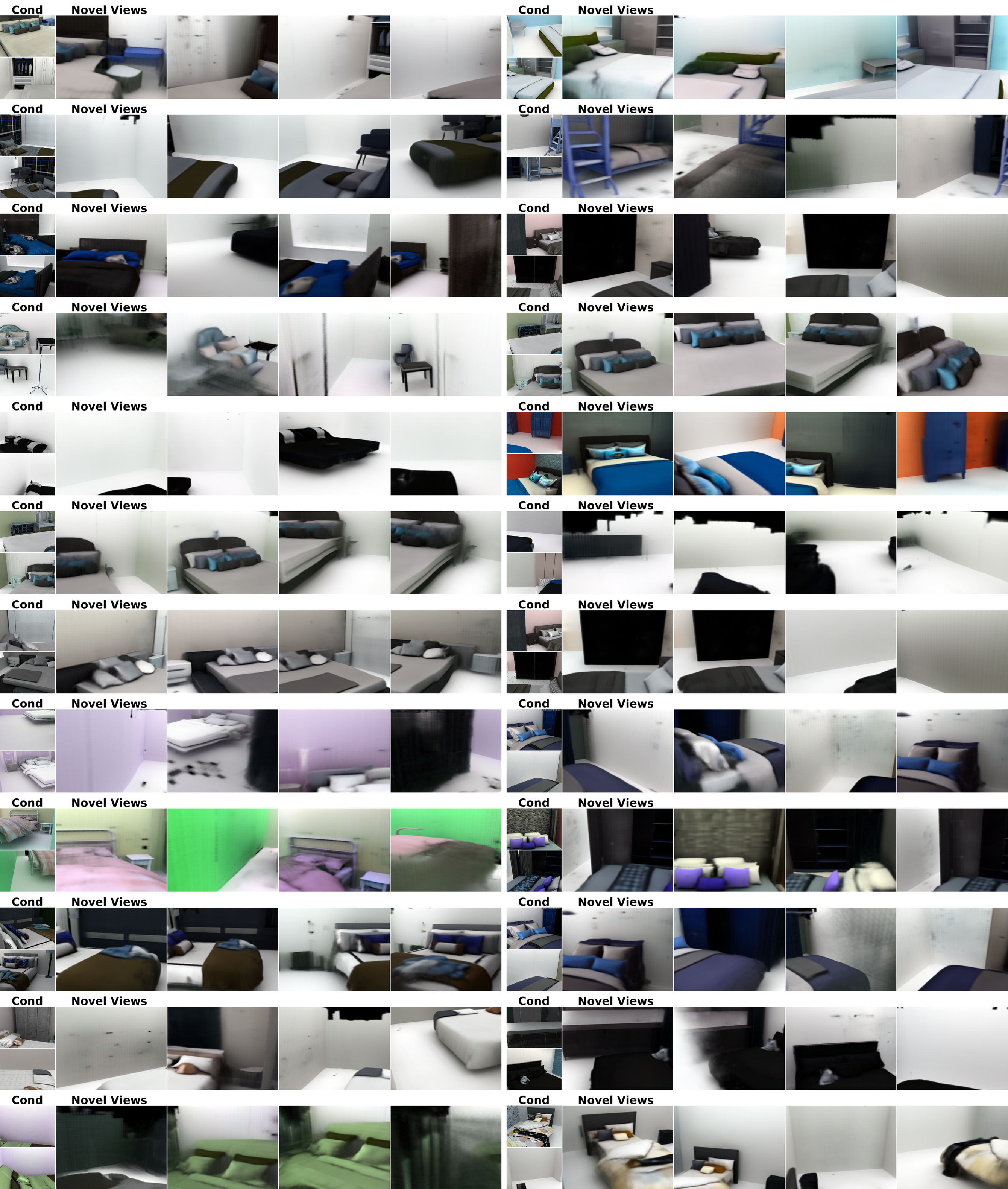} 
  \caption{Novel view synthesis from two conditioning views (left column) on 3D-Front and RealEstate10k datasets.}
    \label{fig:visu_2views}
\end{figure*}

\begin{figure*}[t]
  \centering
  \includegraphics[width=1.0\linewidth]{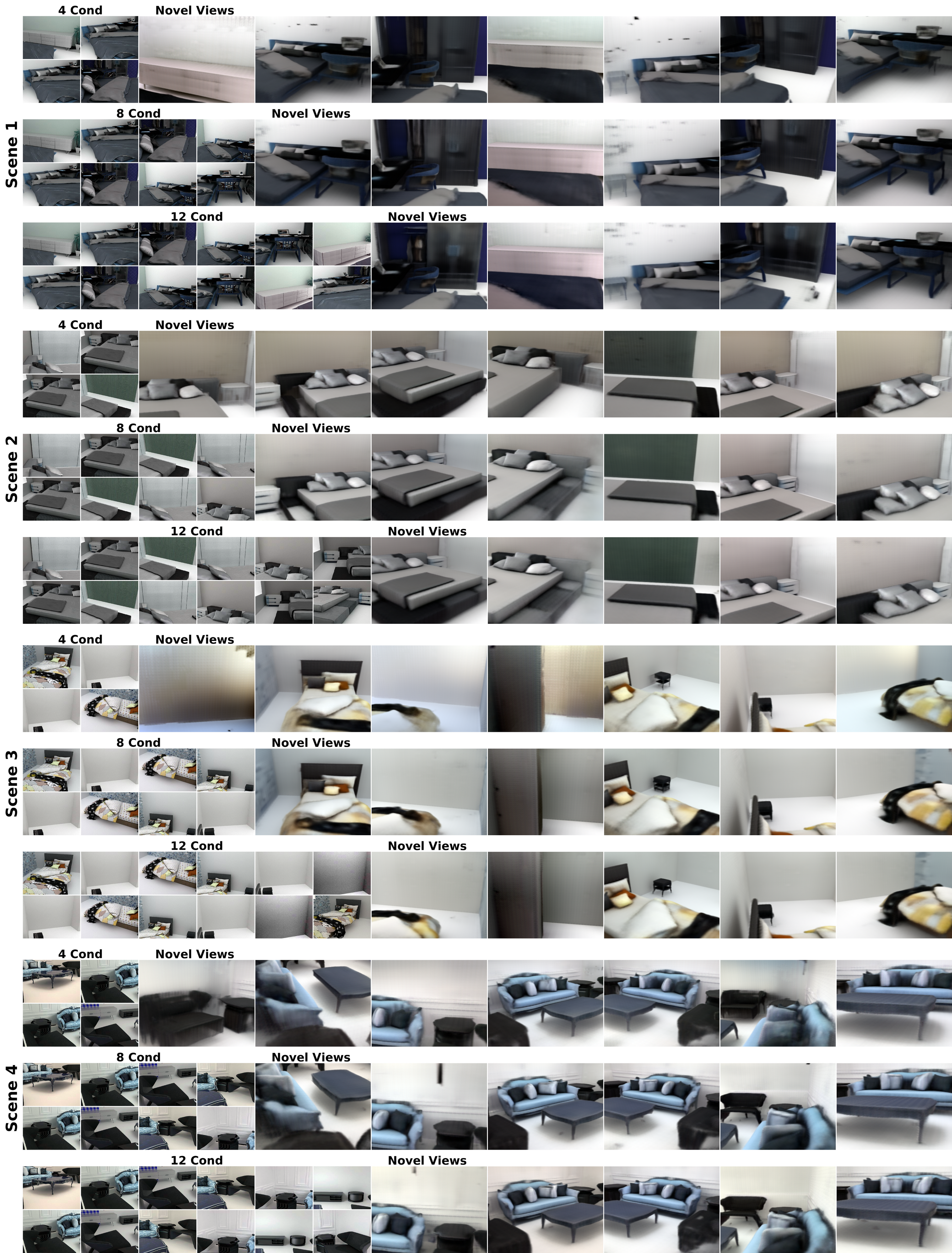} 
  \caption{Comparing view synthesis given different numbers of conditioning views (left: rendered conditioning views, right: rendered novel views). Increasing contextual information leads to less ambiguous scene reconstruction and better renderings. Even in areas without conditioning, \ours is able to sample plausible modes. }
    \label{fig:visu_multicond}
\end{figure*}

In Fig.~\ref{fig:visu_2views}, given two conditioning images (shown in the left column), we display rendered novel views for randomly selected camera viewpoints. \ours is able to extrapolate coherent scene content and geometry far outside the frustums of conditioning cameras. 
On RealEsate10k scenes, cameras exhibit small relative pose variations (both among conditioning views and between conditioning-novel views), thus making novel view synthesis less complicated.  
Additionally, in Fig.~\ref{fig:visu_multicond}, for a single scene, we display rendered novel views from \ours conditioned on a varying number of conditioning views (4/8/12, one set per line). As the number of conditioning views increases, ambiguity decreases and \ours produces more consistent and detailed reconstructions.

\section{Method limitations}
\label{sec:limitation}
Our model relies on voxel-level occupancy predictions to sparsify tensor computations in all components. 
This is computationally efficient, but it has a drawback: if the occupancy model misses voxels inside walls or objects, this creates `holes' in the scene. 
Such holes are visually salient and are currently the most degrading factor for visual quality. Additionally, artifacts such as localized blurriness arise from constraints in voxel resolution and the density of Gaussians allocated per voxel. We hypothesize that these limitations are fundamentally scaling issues; while increasing the dataset size and model capacity would mitigate these artifacts, it would concurrently escalate the computational overhead. One workaround for this is to further refine the output using 2D diffusion models as post-processing. However, this study focuses on establishing the first native 3D generative framework of its kind. We leave the further optimization of the quality-efficiency trade-off to future work.

\end{document}